\documentclass[runningheads]{llncs}
\newcommand{\repeatthanks}{\textsuperscript{\thefootnote}}
\usepackage{subcaption}
\usepackage{comment}
\usepackage{graphicx}
\usepackage{array}
\newcolumntype{C}{>{\centering\arraybackslash}p{6em}}

\usepackage{color}

\begin{document}
%
\title{A Meta-Learning Approach for Medical Image Registration}

%
\titlerunning{A meta-learning approach for medical image registration}
%

\author{Heejung Park\inst{1}\thanks{The authors contribute equally to this paper.} \and
Gyeong Min Lee\inst{1}\repeatthanks \and
Soopil Kim\inst{1} \and
Ga Hyung Ryu\inst{2} \and
Areum Jeong\inst{2} \and
Sang Hyun Park\inst{1}\thanks{Corresponding author}\and
Min Sagong\inst{2}\repeatthanks}


%
\authorrunning{H. Park, G. M. Lee et al.}

%
\institute{Department of Robotics Engineering, DGIST, Daegu, South Korea 
\email{\{qkrgmlwjd39,rud557,soopilkim,shpark13135\}@dgist.ac.kr}\\ \and
Department of Ophthalmology, Yeungnam University College of Medicine,Yeungnam Eye Center, Yeungnam University Hospital, Daegu, South Korea\\
\email{lyugahyung@gmail.com, heart0610@hanmail.net, msagong@ynu.ac.kr}}


%
\maketitle              
\begin{abstract}

Non-rigid registration is a necessary but challenging task in medical imaging studies. Recently, unsupervised registration models have shown good performance, but they often require a large-scale training dataset and long training times. Therefore, in real world application where only dozens to hundreds of image pairs are available, existing models cannot be practically used. To address these limitations, we propose a novel unsupervised registration model which is integrated with a gradient-based meta learning framework. 
In particular, we train a meta learner which finds an initialization point of parameters by utilizing a variety of existing registration datasets. To quickly adapt to various tasks, the meta learner was updated to get close to the center of parameters which are fine-tuned for each registration task. Thereby, our model can adapt to unseen domain tasks via a short fine-tuning process and perform accurate registration. To verify the superiority of our model, we train the model for various 2D medical image registration tasks such as retinal choroid Optical Coherence Tomography Angiography (OCTA), CT organs, and brain MRI scans and test on registration of retinal OCTA Superficial Capillary Plexus (SCP). In our experiments, the proposed model obtained significantly improved performance in terms of accuracy and training time compared to other registration models.

\keywords{Registration \and Unsupervised learning \and Meta learning \and Deep learning.}
\end{abstract}

\section{Introduction}

\begin{figure}[t]
    \centering
    \begin{subfigure}[b]{0.13\textwidth}
        \centering
        \includegraphics[width=\textwidth]{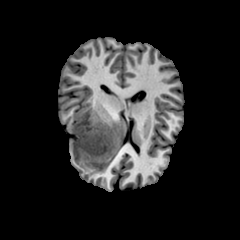}
    \end{subfigure}
    \begin{subfigure}[b]{0.13\textwidth}
        \centering
        \includegraphics[width=\textwidth]{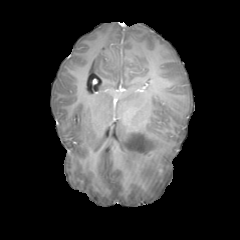}
    \end{subfigure}
    \begin{subfigure}[b]{0.13\textwidth}
        \centering
        \includegraphics[width=\textwidth]{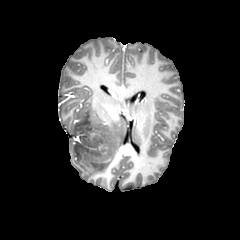}
    \end{subfigure}
    \begin{subfigure}[b]{0.13\textwidth}
        \centering
        \includegraphics[width=\textwidth]{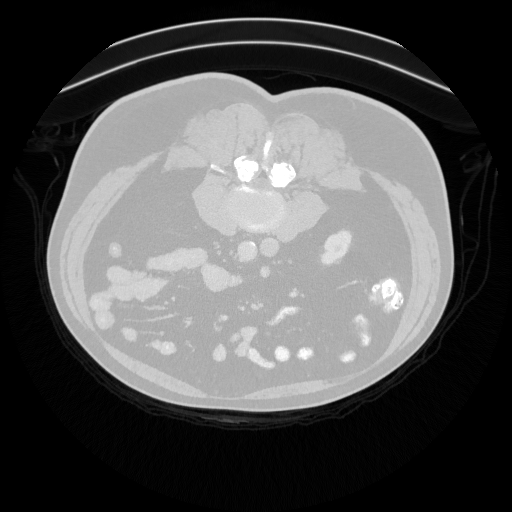}
    \end{subfigure}
    \begin{subfigure}[b]{0.13\textwidth}
        \centering
        \includegraphics[width=\textwidth]{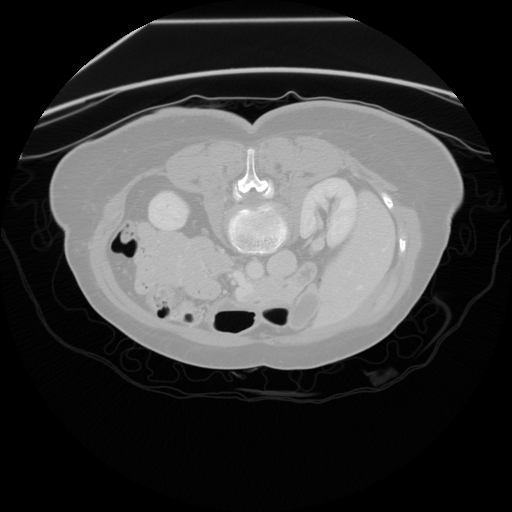}
    \end{subfigure}
    \begin{subfigure}[b]{0.13\textwidth}
        \centering
        \includegraphics[width=\textwidth]{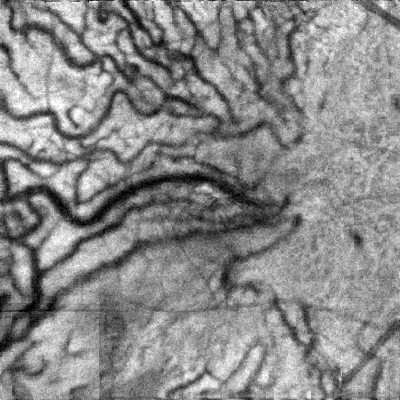}
    \end{subfigure}
    \begin{subfigure}[b]{0.13\textwidth}
        \centering
        \includegraphics[width=\textwidth]{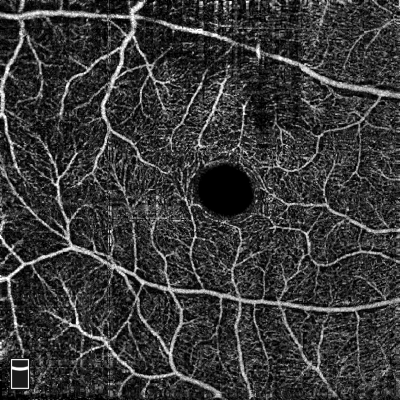}
    \end{subfigure}

    \centering
    
    \begin{subfigure}[b]{0.13\textwidth}
        \centering
        \includegraphics[width=\textwidth]{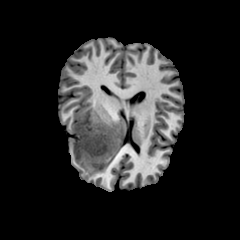}
    \end{subfigure}
    \begin{subfigure}[b]{0.13\textwidth}
        \centering
        \includegraphics[width=\textwidth]{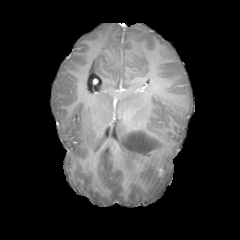}
    \end{subfigure}
    \begin{subfigure}[b]{0.13\textwidth}
        \centering
        \includegraphics[width=\textwidth]{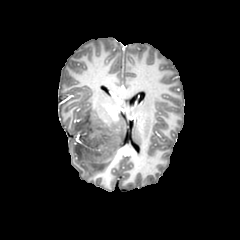}
    \end{subfigure}
    \begin{subfigure}[b]{0.13\textwidth}
        \centering
        \includegraphics[width=\textwidth]{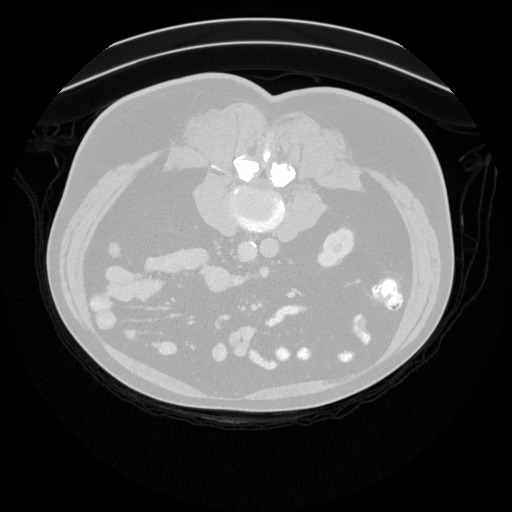}
    \end{subfigure}
    \begin{subfigure}[b]{0.13\textwidth}
        \centering
        \includegraphics[width=\textwidth]{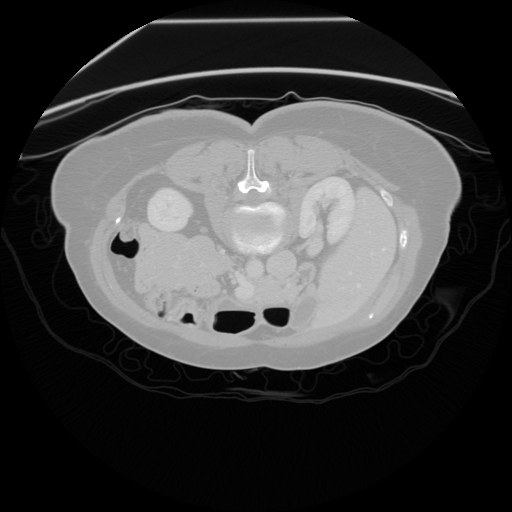}
    \end{subfigure}
    \begin{subfigure}[b]{0.13\textwidth}
        \centering
        \includegraphics[width=\textwidth]{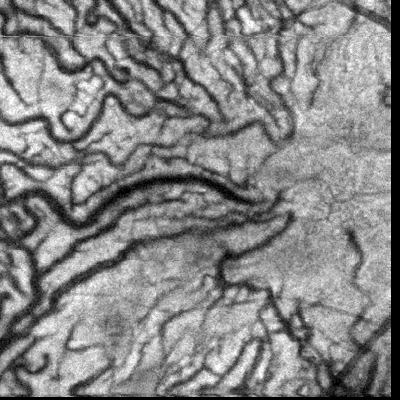}
    \end{subfigure}
    \begin{subfigure}[b]{0.13\textwidth}
        \centering
        \includegraphics[width=\textwidth]{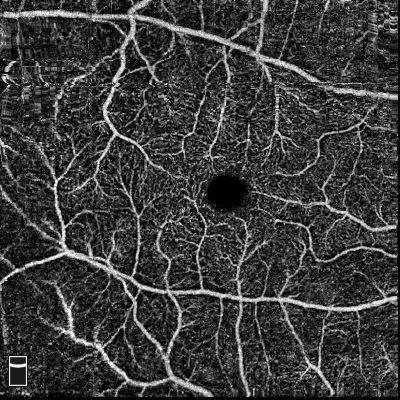}
    \end{subfigure}

    \caption{Example of moving (top) and fixed (bottom) image pairs. From left to right: brain MRI (T2w, T1w, T1Gd), abdomen CT from different subjects, retinal OCTA choroid, and retinal OCTA SCP scans}
\label{fig:sampledata}
\end{figure}

Non-rigid registration is necessary to quantify image changes over time or between different patients. Recently, unsupervised deep learning-based registration methods \cite{Li2017,Vos2017,Vos2019,Balakrishnan2019} that do not require ground truth correspondences have been proposed and achieved state of the art performance in many applications. However, most methods addressed registration of images from a single domain with a large number of training data, thus performance can greatly reduce when the model trained with image pairs in a different domain is applied to a new image pair. Moreover, convergence speed is often slow in model updates for registering new data since image characteristics of the new data may be different to that of the data used in the pre-training stage.



Several studies have been conducted to address this problem. For example, Guan et al.\cite{guan2020RegTransfer} applied transfer learning to medical image registration by using data from multiple patients. However, this was proposed for a supervised registration training setting and could not utilize data from other domains. Recently, meta-learning have been proposed to learn generalizable knowledge and adapt fast to unseen tasks via task-level training. Wang et al.\cite{wang20203d} proposed a meta-learning technique that can address 3D point cloud registration by estimating convolution weights for unseen data registration from external meta modules. However, their model is trained in a supervised manner using ground truth correspondences, and thus cannot be applied to medical image registration where it is difficult to make ground truth correspondences.


To address this problem, we propose a meta-learning-based registration meth-od that can efficiently use data from different domains. Specifically, we integrate an unsupervised learning-based registration model in a gradient-based meta-learning framework. The registration model is first trained using multiple registration datasets and then performs task-level learning using the multi-task data. The meta-learner finds an initialization point which can quickly adapt to various registration problems. After that, fine tuning is performed with the data of the target domain and applied to the test set. To demonstrate superiority, we trained the model for various 2D registration tasks from retinal Optical Coherence Tomography Angiography (OCTA) choroid, abdomen CT, and brain MRI scans and tested it on a registration of retinal OCTA Superficial Capillary Plexus (SCP) scans as shown in Fig. \ref{fig:sampledata}.

The main contributions of this work are summarized as follows: (1) We propose a novel registration framework which quickly adapts to new tasks on unseen domains via meta-learning framework. 
(2) Our framework can effectively utilize existing registration data to learn a new registration task from unseen domains. Lastly, (3) any registration models can be integrated in our gradient-based meta-learning framework.

\subsection{Related works}
\noindent{\textbf{Registration via unsupervised deep learning model:}}
Recently, unsupervised learning models have been proposed for registration to alleviate the effort of making ground truth correspondence. VoxelMorph\cite{Balakrishnan2019} directly predicts the deformation map and  deforms the moving image using a spatial transformer\cite{Jaderberg2015} which can deform the moving image using deformation maps. The spatial transformer makes the network enable backpropagation in optimization so that the similarity between deformed moving image and target image is maximized. Based on this model, Zhao et al.\cite{zhao2019recursive} proposed a cascaded framework to address large displacements by recursively deforming the images. Mahapatra et al.\cite{mahapatra2018deformable} further utilize a GAN architecture to improve the registration performance, while Lee et al.\cite{lee2020unsupervised} utilize additional features to perform the registration between images with different characteristics. Notably, it is hard to apply these methods to unseen domains since they focused on the registration in a single domain. We propose a meta-learning-based non-rigid registration model which can quickly adapt to unseen domain.

\noindent{\textbf{Gradient based meta learning framework:}} 
Gradient based meta learning was recently proposed to address limitations of supervised learning that depends on a large training dataset. In particular, MAML\cite{finn2017MAML} finds an initialization point that can quickly adapt to various tasks with only a few data samples. It was applied to various tasks such as few shot segmentation\cite{liu2020shape,hendryx2019metaSEG} and zero shot super resolution\cite{soh2020metaZSSR}. Based on this model, several methods were proposed to improve MAML by adding learnable parameters for updating direction and learning rate\cite{li2017metaSGD}, dividing training modules to use deeper layers \cite{sun2019metaTransfer}, and simplifying computation processes\cite{nichol2018FOMAML} respectively. Especially, Nichol et al.\cite{nichol2018FOMAML} proposed FOMAML and Reptile to reduce the computation of MAML by ignoring second order derivative terms and reformulate the meta learning framework. While most meta-learning frameworks require train/test separated data for each task, Reptile is similar to ordinary gradient-based learning framework. Thus, we adopt a Reptile-based meta-learning model to learn generalizable registration knowledge from various tasks.

\section{Method}

\begin{figure}[!t]
\includegraphics[width=1.0\textwidth]{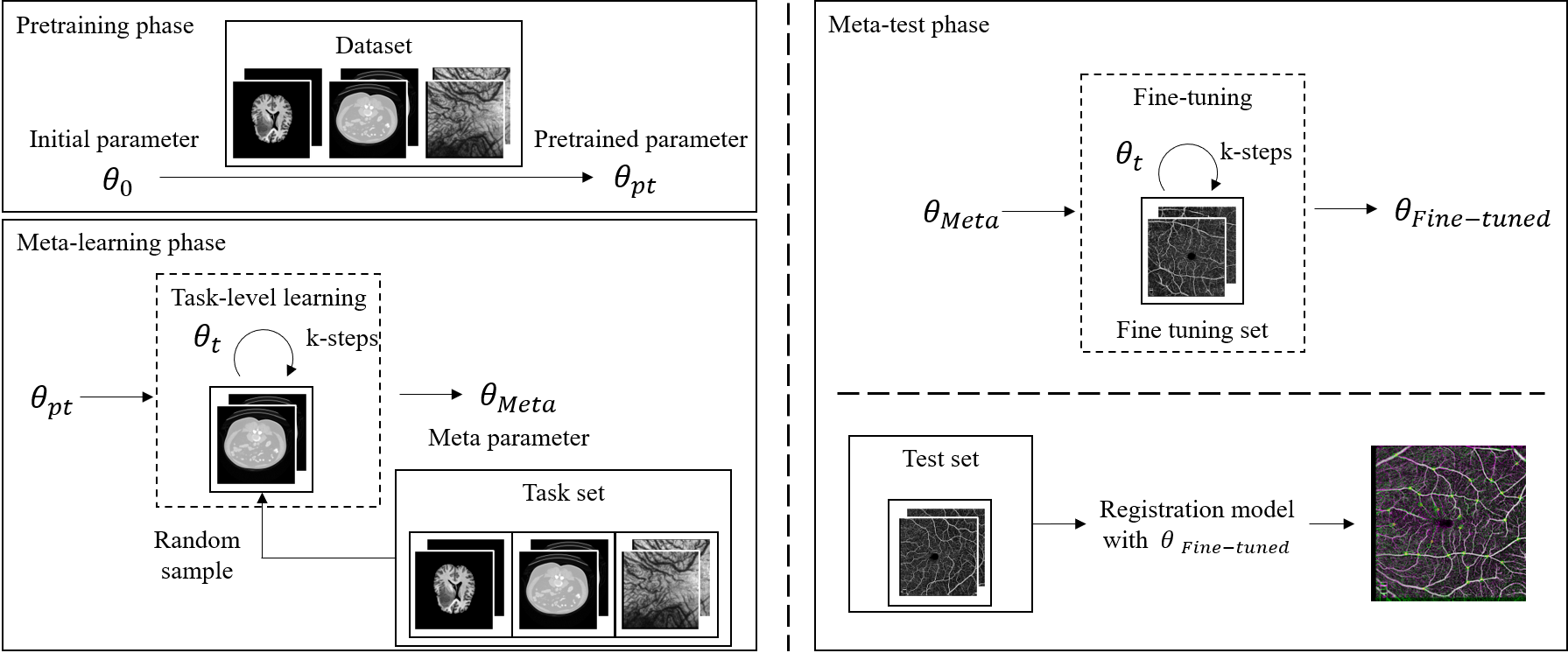}
\centering
\caption{Overview of the proposed method.}
\label{fig::Propose_net_overall}
\end{figure}

The proposed meta learning framework is shown in Fig.~\ref{fig::Propose_net_overall}. Unlike the conventional registration methods that only use data from the target domain $D_{target}$, we utilize data from multiple domains $D_{source}$. First, an unsupervised deformable registration model $G_{\theta}$ is pretrained with $D_{source}$ to learn optimal parameters $\theta$ to predict a displacement map $\phi$ for moving and fixed image pairs $(M,F)$. Then, $G_{\theta_{meta}}$ is trained with Reptile\cite{nichol2018FOMAML}, a gradient based meta learning algorithm, using the same dataset. Unlike the pretraining step, we create several task sets $T_{source}=T_1, T_2, ...$ from $D_{source}$ to perform task-level training. Meta knowledge is learned from this and then applied to $D_{target}$. For testing, $G_{\theta_{meta}}$ is fine-tuned to target data $T_{fine-tune}$ from $D_{target}$ and used to perform the registration on testing data $T_{test}$.

\subsubsection{Pretraining of registration model using Unsupervised learning}
Our meta registration model is compatible with any registration models. In our experiment, we used VoxelMorph\cite{Balakrishnan2019} as the registration model; a U-Net\cite{Ronneberger2015} style encoder-decoder with a spatial transformer. $M$ and $F$ were concatenated in the channel direction and used as inputs for $G_\theta$. The deformation field $\phi$ is  obtained as a result of $G(M,F)$ and used to make a deformed moving image $M(\phi)$ using the spatial transformer. For optimization, we employ a loss based on the summation of the similarity between $M(\phi)$ and $F$ including the smoothness of $\phi$. Formally:
\begin{equation}
L(F,M,\phi) = -CC(F,M(\phi)) + \lambda \sum  \Vert\bigtriangledown \phi \Vert^{2},
\label{total_loss}
\end{equation}
where $\lambda$ is a gradient regularization parameters, $CC(F,M(\phi))$ is the cross-correlation (CC) in $9\times9$ window between $F$ and $M(\phi)$, and $\sum  \Vert\bigtriangledown \phi \Vert^{2}$ penalizes the magnitude of $\phi$ gradients.

Prior to meta learning stage, $G_\theta$ is pretrained with $D_{source}$. Registration problems from different domains can be considered as a single domain problem to find representative features from $M$ and $F$ pairs and compute $\phi$. By learning basic representations, we can boost the meta learning process which is sometimes unstable to learn meta knowledge when training from scratch. After loss convergence, the pre-trained model $\theta_{pt}$ is saved.

\subsubsection{Model update via Meta-learning Framework}
After pretraining $G_{\theta_{pt}}$, the model is trained with meta learning before fine tuning to an unseen registration problem. Among several gradient based meta learning algorithms, we adopted Reptile\cite{nichol2018FOMAML}, which can be easily integrated with the registration models. With this framework, separation of support(training) and query(test) data is not necessary for each dataset, and thus it can be applied to unsupervised learning.

In the meta learning process, the model learns parameter $\theta_{meta}$ according to following steps: When batch size is $m$, we randomly sample $m$ tasks from $T_{source}$ and then $(M,F)$ pair is randomly selected from each task. $\theta_{meta}$ is updated using unsupervised registration loss following Eq. (\ref{total_loss}). After updating parameters $k$ times for each task, we obtain fine-tuned parameter $\theta_t$ corresponding to each task. Regarding $(\theta_t - \theta_{meta})$ as a gradient, $\theta_{meta}$ is updated as:
\begin{equation}
\theta_{meta} = \theta_{meta} + \alpha \frac{1}{m} \sum_{t=1}^m (\theta_t - \theta_{meta}),
\end{equation}
where $\alpha$ is learning rate. Through the corresponding process, $\theta_{meta}$ converges to a certain parameter which can quickly adapt to various registration tasks.

In the test stage, $G_\theta$ adapts to the target task using $T_{fine-tune}$ from obtained parameter $\theta_{meta}$. After fast adaptation, the model with fine-tuned parameters can perform accurate registrations on test data $T_{test}$ without additional training.

\subsubsection{Implementation Details} 
We implemented the proposed method using pytorch\cite{NEURIPS2019} on Intel i9-9900K CPU,NVIDIA GeForce RTX 2080 Ti with 64 GB RAM. To update $\theta_t$ and $\theta_{meta}$, two Adam~\cite{Kingma2014} optimizers were used and learning rate of both model were set to $1e^{-4}$. In each meta-learning step, a batch consisted of randomly sampled 3 tasks and the number of update k was 10. We used the same gradient regularization parameter $\lambda$ set to 1 for all methods.
\section{Experimental Results}

\subsubsection{Dataset} 
In our experiment, we used four datasets including retinal OCTA SCP, retinal OCTA choroid, abdomen CT, and Brain MRI. Both OCTA SCP and choroid datasets contained 368 moving and fixed image pairs collected from local university hospital, some of which were taken from same subjects at different times. The abdomen CT and brain MRI images were obtained from public Decathlon dataset~\cite{simpson2019large}. Here, we define three tasks according to modality (T1w, T1Gd, and T2w) from the brain MRI dataset and two tasks in the abdomen CT dataset. Each 3D volume was divided in multiple axial slices and adjacent two slices were defined as a $(M,F)$ pair. All images were resized to a size of $400\times400$ and histogram equalization was applied. Also, the range of intensity was rescaled to [0,1]. For training, we defined a set of five tasks as the source task set $T_{source} =$ $T_{brain T1}$, $T_{brain T1Gd}$, $T_{brain T2}$, $T_{abdomen}$, $T_{Choroid}$. For evaluation, retinal OCTA SCP dataset was used as target domain data $D_{target}$. It was divided into a fine-tuning set $T_{fine-tune}$ (294 pairs) and a test set $T_{test}$ (74 pairs). For evaluation, we manually labeled 20$\sim$30 bifurcation points on image pairs in $T_{test}$.


\subsubsection{Experimental settings}
To evaluate the registration performance, we compared our method to four models (\textit{Model-Seen, Model-Unseen, Transfer}, and \textit{Fine-tune}) with different training strategy. All methods were tested on $T_{test}$. \textit{Model-Seen} and \textit{Model-Unseen} were both trained only with target domain data. \textit{Model-Seen} was trained using both $T_{fine-tune}$ and $T_{test}$, while \textit{Model-Unseen} was trained only with $T_{fine-tune}$. \textit{Transfer} and \textit{Fine-tune} models were transfer learning-based models. \textit{Transfer} was trained with $T_{source}$ and then applied to $T_{test}$ without additional update. Meanwhile, \textit{Fine-tune} was additionally trained from the parameters of \textit{Transfer} using $T_{fine-tune}$ and then applied to $T_{test}$. Our proposed model performed the task-level training from the parameters of \textit{Transfer}. It was trained with $T_{source}$, and then fine-tuned to $T_{fine-tune}$. We tested these models after their loss converges, i.e., 25k, 38k, 6k, and 23k epochs for \textit{Model-Seen}, \textit{Model-Unseen}, pre-training and meta-iterations, respectively. As evaluation metrics, we measured average pixel distances between deformed bifurcation points on $M(\phi)$ and correspondences on $F$ and the normalized cross-correlation (NCC) between $M(\phi)$ and $F$.

\begin{table}[!t] \centering
	\caption{Performance comparison of the proposed model against the other baselines on OCTA SCP dataset. (Pixel distance $\pm$ std and NCC $\pm$ std)}\label{result_table1}
	\begin{tabular}{p{0.3\textwidth}>{\centering}p{0.2\textwidth}>{\centering}p{0.2\textwidth}>{\centering}p{0.1\textwidth}>{\centering\arraybackslash}p{0.1\textwidth}}
		\hline
		Method & Distance & NCC
		\tabularnewline
		\hline
		Not deformed & 10.07 $\pm$ 3.99 & 0.06 $\pm$ 0.02
		\tabularnewline
		
		%
		
		\hline
		\textit{Model-Seen} & 5.55 $\pm$ 3.99 & 0.24 $\pm$ 0.09
		\tabularnewline
		
		\hline
		\textit{Model-Unseen} & 5.85 $\pm$ 3.99 & 0.22 $\pm$ 0.08
		\tabularnewline
		
		\hline
		\textit{Transfer} & 6.19 $\pm$ 3.59 & 0.16 $\pm$ 0.08
		\tabularnewline
		
		\hline
		\textit{Fine-tune} - 10 epochs & 5.90 $\pm$ 3.66 & 0.18 $\pm$ 0.08
		\tabularnewline
		
		\hline
		\textit{Fine-tune} - 2000 epochs & 4.81 $\pm$ 2.93 & 0.24 $\pm$ 0.08
		\tabularnewline
		
		\hline
		Ours - 10 epochs & 5.05 $\pm$ 3.27 & 0.23 $\pm$ 0.09
		\tabularnewline
		
		\hline
		Ours - 2000 epochs. & 4.52 $\pm$ 2.93 & 0.28 $\pm$ 0.10
		\tabularnewline

		\hline
	\end{tabular}
\end{table}

\begin{figure}[!t]
	\centering
	\begin{subfigure}[b]{0.32\textwidth}
		\centering
		\includegraphics[width=\textwidth]{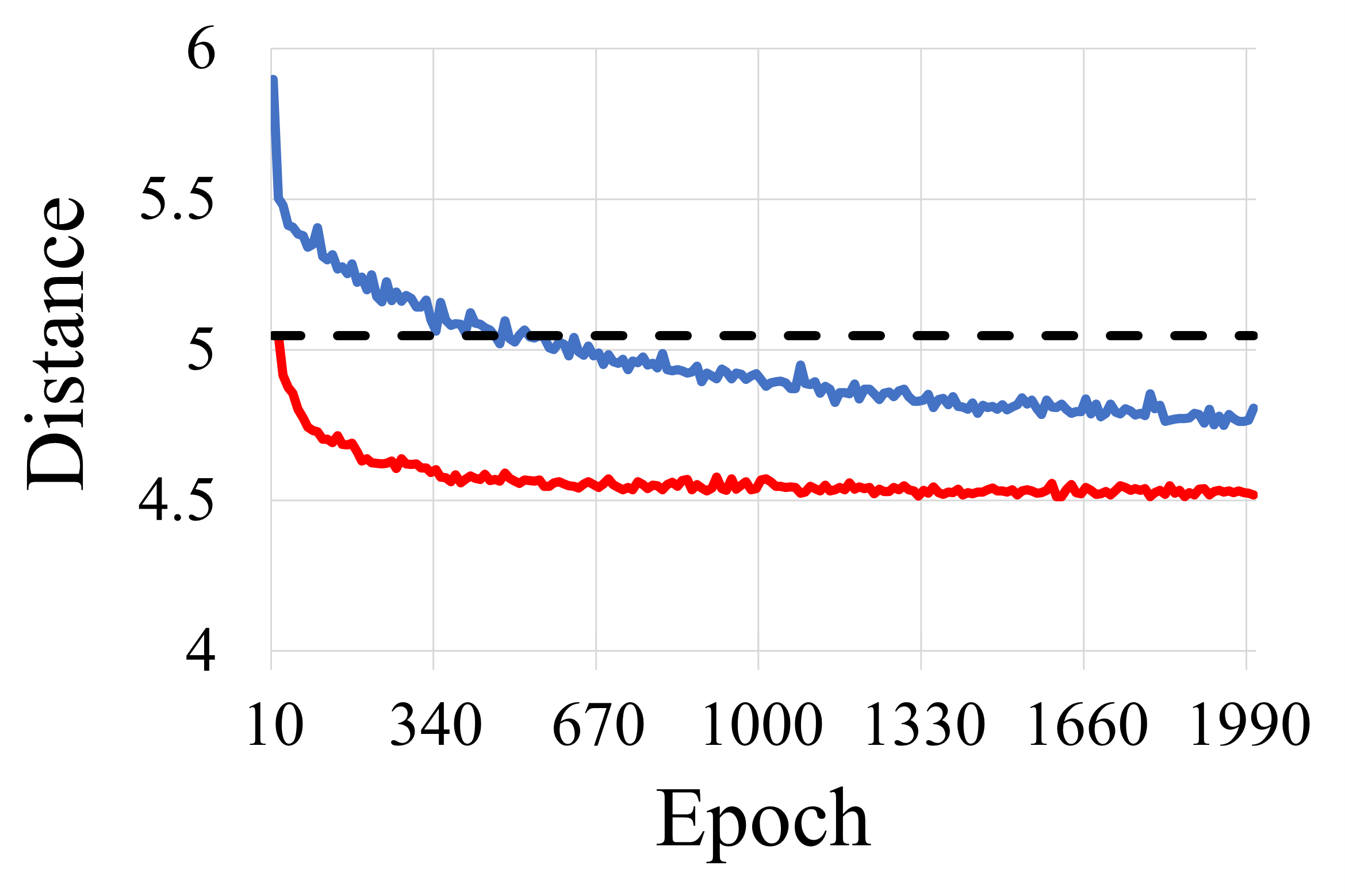}
		\caption{}
	\end{subfigure}
	\begin{subfigure}[b]{0.32\textwidth}
		\centering
		\includegraphics[width=\textwidth]{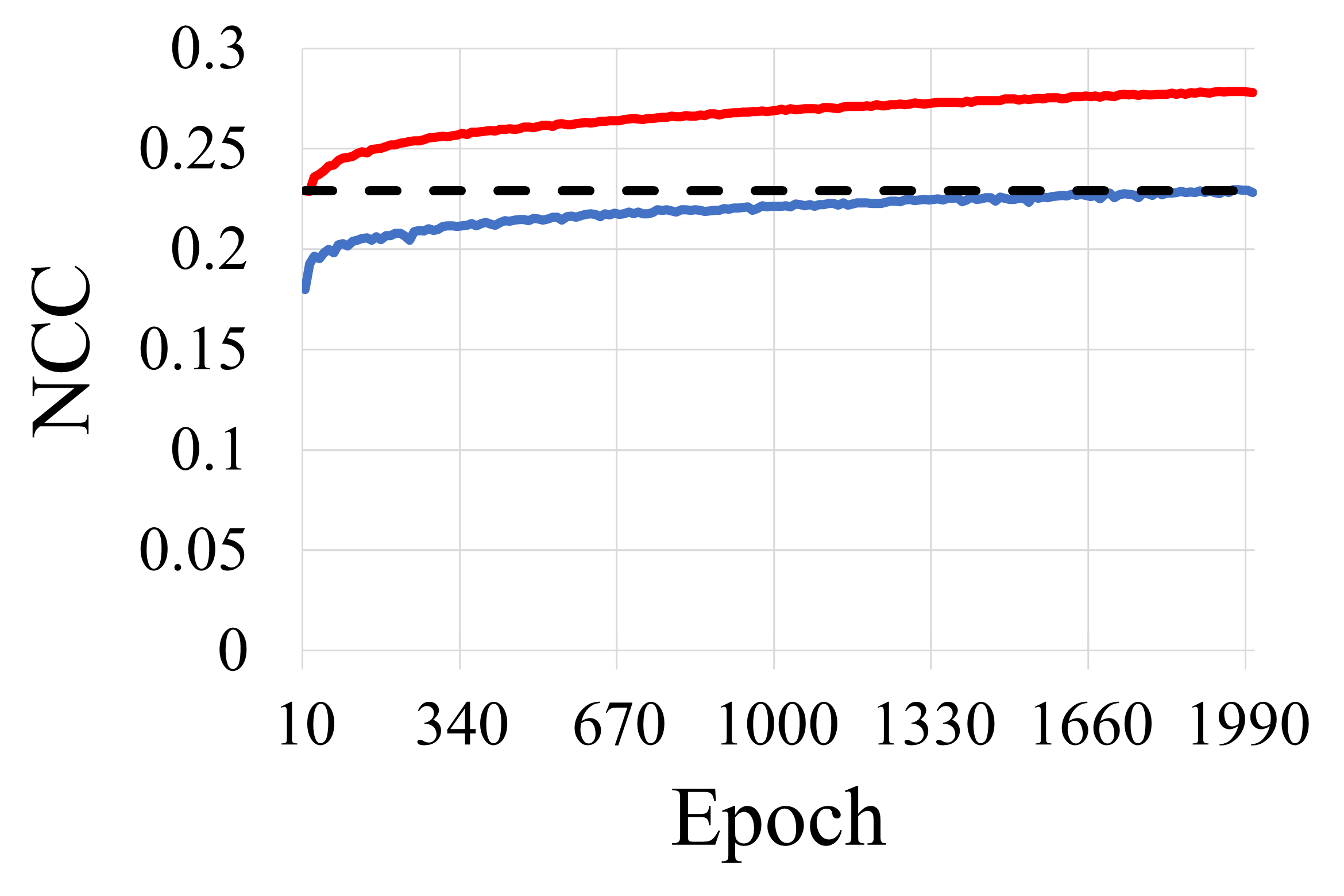}
		\caption{}
	\end{subfigure}
	\begin{subfigure}[b]{0.32\textwidth}
		\centering
		\includegraphics[width=\textwidth]{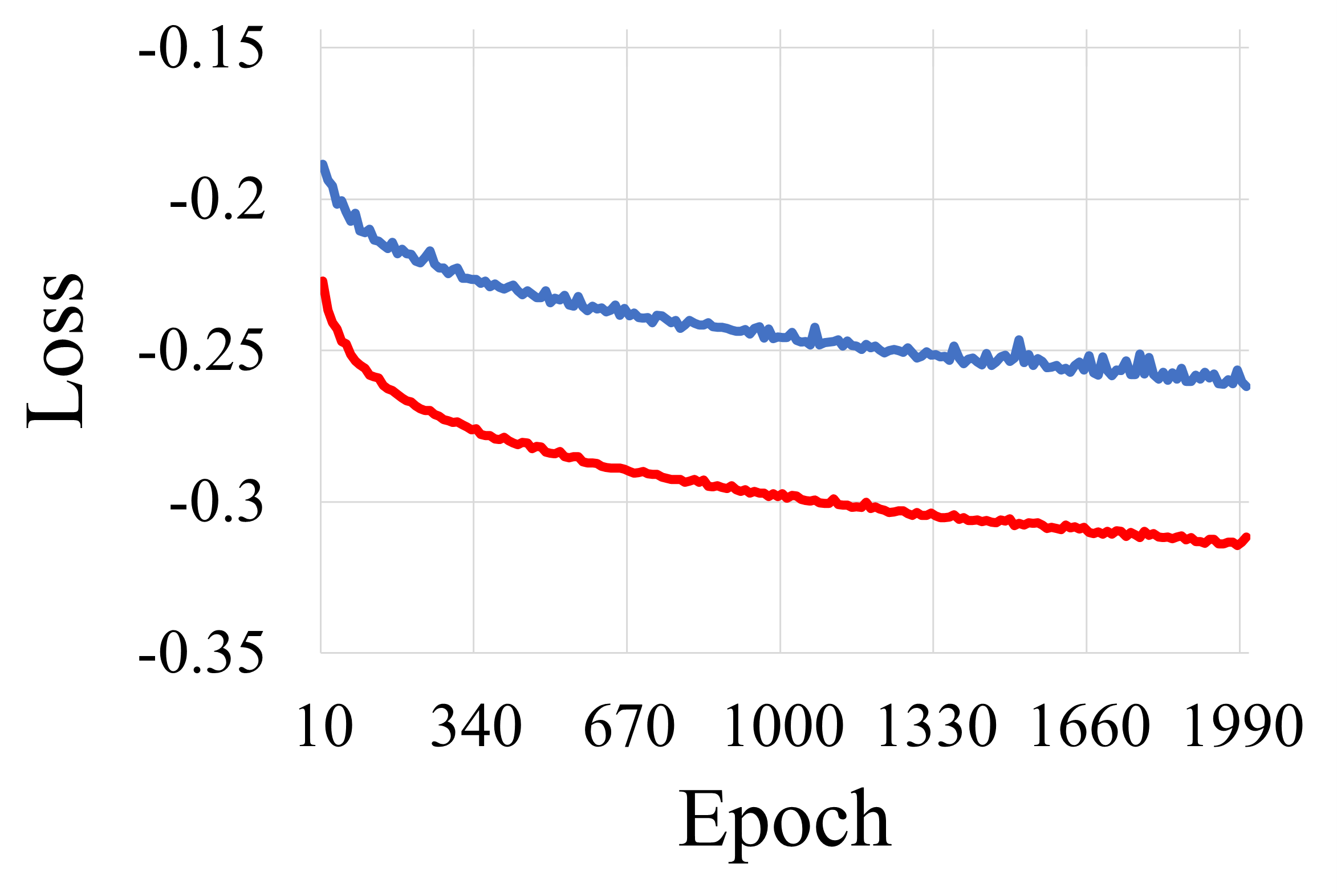}
		\caption{}
	\end{subfigure}
	
	\caption{Comparison of fine-tuning phase between \textit{Fine-tune} (solid blue) and Ours (solid red) with Ours-10 epoch performance (dotted black). (a) pixel distance, (b) NCC, and (c) training loss were recorded in every 10 epochs from 10 to 2000 epochs.} 
	\label{fig:result_graph}
\end{figure}

\subsubsection{Quantitative Results}
We present the average and standard deviation of pixel distance between matching points and NCC of comparison methods in Table \ref{result_table1}. 
\textit{Model-Seen} and \textit{Model-Unseen} outperformed \textit{Transfer} method since the brain MRI and abdomen CT images used for training have significantly different characteristics from OCTA SCP. Even so, using data from other domains for training may affect the registration performance of the target domain, and showed better performance than \textit{Model-Seen} and \textit{Model-Unseen} through fine tuning. The models with fine-tuning of 10 epochs showed similar performance to \textit{Model-Unseen} and with fine tuning of 2000 epochs showed better performance than \textit{Model-Seen}.

\begin{figure}[!t]
	\centering
	\begin{subfigure}[b]{0.15\textwidth}
		\centering
		\includegraphics[width=\textwidth]{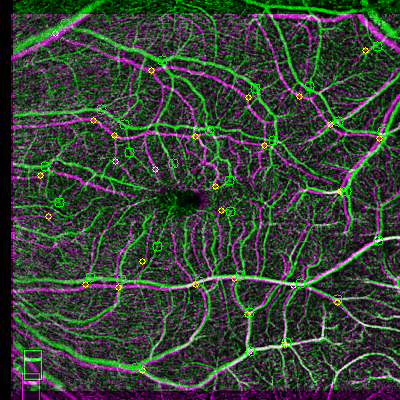}
	\end{subfigure}
	\begin{subfigure}[b]{0.15\textwidth}
		\centering
		\includegraphics[width=\textwidth]{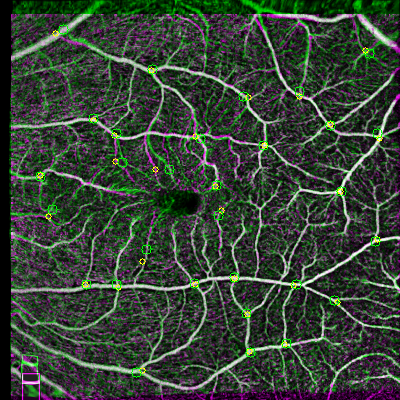}
	\end{subfigure}
	\begin{subfigure}[b]{0.15\textwidth}
		\centering
		\includegraphics[width=\textwidth]{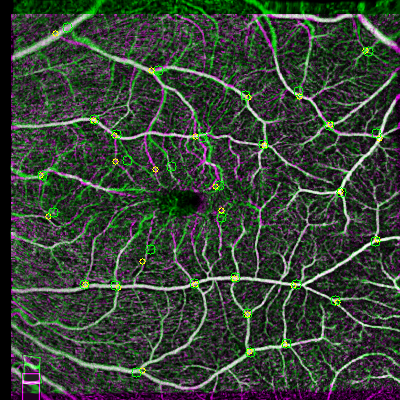}
	\end{subfigure}
	\begin{subfigure}[b]{0.15\textwidth}
		\centering
		\includegraphics[width=\textwidth]{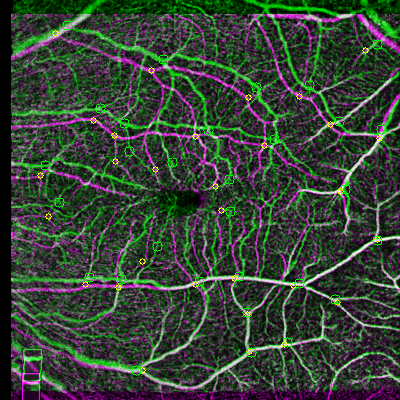}
	\end{subfigure}
	\begin{subfigure}[b]{0.15\textwidth}
		\centering
		\includegraphics[width=\textwidth]{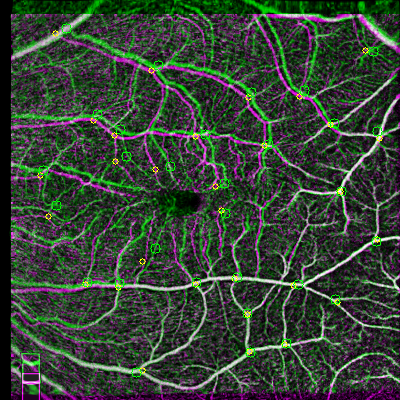}
	\end{subfigure}
	\begin{subfigure}[b]{0.15\textwidth}
		\centering
		\includegraphics[width=\textwidth]{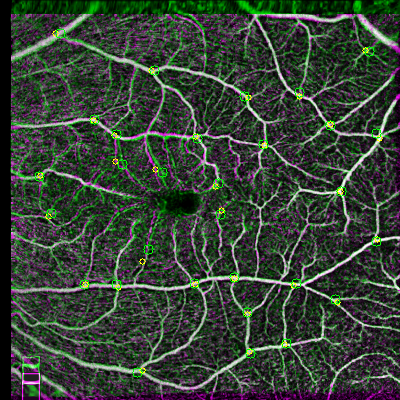}
	\end{subfigure}
	
	\centering
	\begin{subfigure}[b]{0.15\textwidth}
		\centering
		\includegraphics[width=\textwidth]{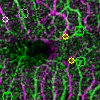}
	\end{subfigure}
	\begin{subfigure}[b]{0.15\textwidth}
		\centering
		\includegraphics[width=\textwidth]{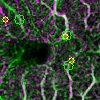}
	\end{subfigure}
	\begin{subfigure}[b]{0.15\textwidth}
		\centering
		\includegraphics[width=\textwidth]{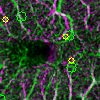}
	\end{subfigure}
	\begin{subfigure}[b]{0.15\textwidth}
		\centering
		\includegraphics[width=\textwidth]{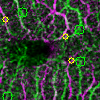}
	\end{subfigure}
	\begin{subfigure}[b]{0.15\textwidth}
		\centering
		\includegraphics[width=\textwidth]{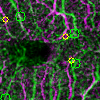}
	\end{subfigure}
	\begin{subfigure}[b]{0.15\textwidth}
		\centering
		\includegraphics[width=\textwidth]{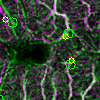}
	\end{subfigure}
	
	\hfill
	
	\centering
	\begin{subfigure}[b]{0.15\textwidth}
		\centering
		\includegraphics[width=\textwidth]{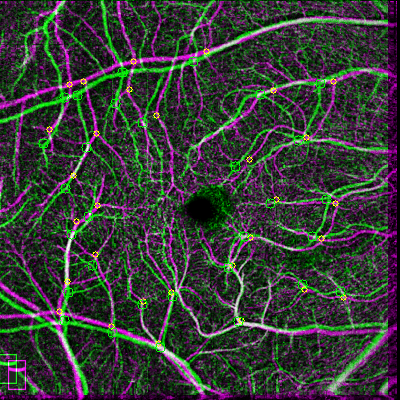}
	\end{subfigure}
	\begin{subfigure}[b]{0.15\textwidth}
		\centering
		\includegraphics[width=\textwidth]{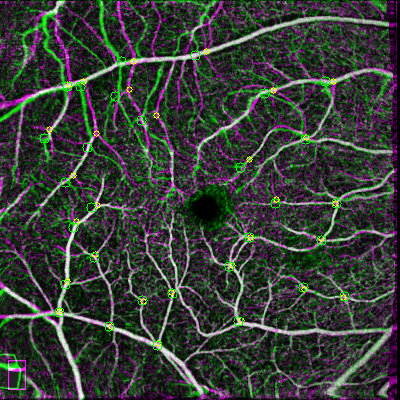}
	\end{subfigure}
	\begin{subfigure}[b]{0.15\textwidth}
		\centering
		\includegraphics[width=\textwidth]{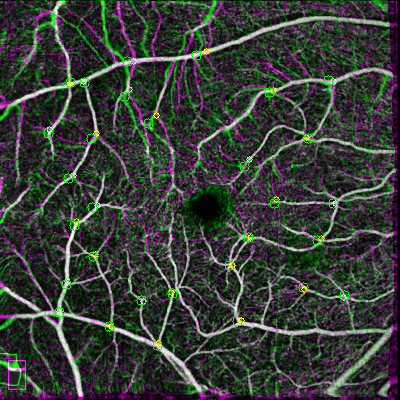}
	\end{subfigure}
	\begin{subfigure}[b]{0.15\textwidth}
		\centering
		\includegraphics[width=\textwidth]{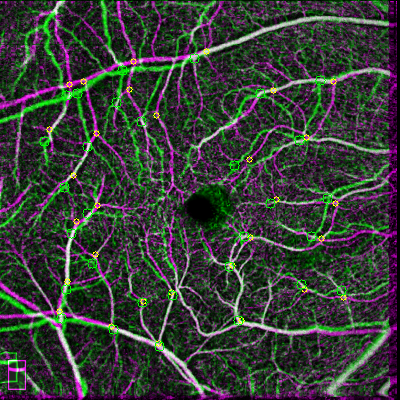}
	\end{subfigure}
	\begin{subfigure}[b]{0.15\textwidth}
		\centering
		\includegraphics[width=\textwidth]{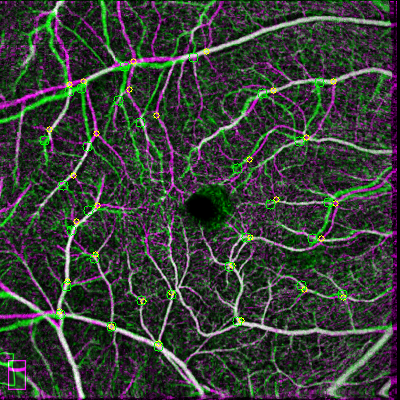}
	\end{subfigure}
	\begin{subfigure}[b]{0.15\textwidth}
		\centering
		\includegraphics[width=\textwidth]{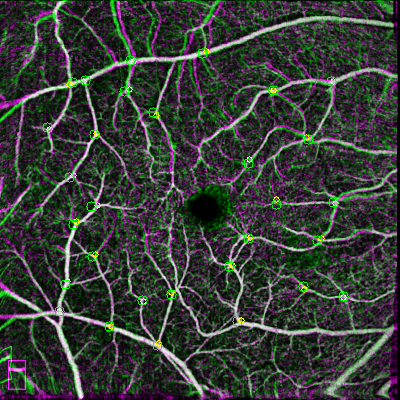}
	\end{subfigure}
	
	\centering
	\begin{subfigure}[b]{0.15\textwidth}
		\centering
		\includegraphics[width=\textwidth]{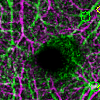}
		\caption{}
	\end{subfigure}
	\begin{subfigure}[b]{0.15\textwidth}
		\centering
		\includegraphics[width=\textwidth]{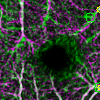}
		\caption{}
	\end{subfigure}
	\begin{subfigure}[b]{0.15\textwidth}
		\centering
		\includegraphics[width=\textwidth]{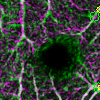}
		\caption{}
	\end{subfigure}
	\begin{subfigure}[b]{0.15\textwidth}
		\centering
		\includegraphics[width=\textwidth]{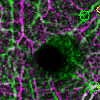}
		\caption{}
	\end{subfigure}
	\begin{subfigure}[b]{0.15\textwidth}
		\centering
		\includegraphics[width=\textwidth]{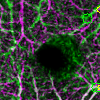}
		\caption{}
	\end{subfigure}
	\begin{subfigure}[b]{0.15\textwidth}
		\centering
		\includegraphics[width=\textwidth]{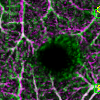}
		\caption{}
	\end{subfigure}
	
	\caption{Qualitative results. Each image shows an overlap of $M(\phi)$ (purple) and $F$ (green) with corresponding points in deformed points (green) and fixed points (yellow). Overlapping area appears in white. (a) Not deformed, (b) \textit{Model-Seen}, (c) \textit{Model-Unseen}, (d) \textit{Transfer}, (e) \textit{Fine-tune} - 10epochs, (f) Ours - 10epochs}
	\label{fig:result_images}
\end{figure}

The proposed meta-learning method efficiently learns the data of the source domain and showed better performance than the simple fine tuning based models. Ours with 10 epochs update showed better performance than \textit{Model-Seen} and ours with 2000 epochs significantly outperformed \textit{Fine-tune} – 2000 epochs.


When we compared the fine tuning stage between proposed method and \textit{Fine-tune}, we observed similar tendency from (a) distance, (b) NCC and (c) loss graphs as shown in Fig. \ref{fig:result_graph}. Even though \textit{Fine-tune} could achieve the initial performance of our model after a long training time, i.e. more than 500 in distance graph and 1500 epochs in NCC graph, respectively, our model achieved better performance even after saturation. In addition, the proposed method requires only about 600 epochs to converge, compared to \textit{Fine-tune} which requires about 1500 epochs. These results show that our proposed method is updated from a better initialization point to learn unseen tasks fast. 


\subsubsection{Qualitative Results}
Fig. \ref{fig:result_images} shows qualitative results of comparison methods. Our qualitative observations are consistent with quantitative comparisons. \textit{Transfer} predicted poor registration results since the fine-tuning was not considered. When the model is fine-tuned, there was improvement as shown in \textit{Fine-tune}-10epochs, but there were gaps between the corresponding points and vessels. \textit{Model-Seen} predicted relatively better results, but some details were still not registered correctly. On the other hand, our model obtained closer distance between corresponding points and well-aligned vessels.


%
\section{Conclusions}
In this paper, we propose a novel registration model which is adaptable to unseen tasks. As a first application of meta learning to unsuerpvised registration task, our model effectively utilizes existing multi-domain data to learn an initialization which can adapt to various tasks fast. Our experiments on different registration approaches showed superiority of our proposed method. In future work, we will develop a meta learning model that can work on a more challenging task such as cross-domain registration.

%

\bibliographystyle{splncs04}
\bibliography{bibliography}

\end{document}